\documentclass[conference]{IEEEtran}
\IEEEoverridecommandlockouts
\usepackage{cite}

\usepackage[left=1.62cm,right=1.62cm,top=1.9cm]{geometry}
\usepackage{xcolor}
\usepackage{graphicx}
\usepackage{algorithm}
\usepackage{algpseudocode}
\usepackage{amsmath}

\usepackage{tabularray}
\usepackage{caption}
\usepackage{subcaption}

\usepackage{comment}

\newsavebox\mybox

\def\BibTeX{{\rm B\kern-.05em{\sc i\kern-.025em b}\kern-.08em
    T\kern-.1667em\lower.7ex\hbox{E}\kern-.125emX}}
\begin{document}

\title{
ConvLSTMTransNet: A Hybrid Deep Learning Approach for Internet Traffic Telemetry

}

\author{Sajal~Saha$^{1}$, Saikat Das$^{2}$, and         
        Glaucio~H.S.~Carvalho$^{3}$,\\ 
       $^1$Department of Computer Science, University of Northern British Columbia, Prince George, BC, Canada.\\
       $^2$Department of Computer Science, Utah Valley University, Orem, UT, USA.\\
       $^3$Department of Computer Science and Engineering, Brock University, St. Catharines, ON, Canada.\\
       \textit{Email:} \textit{sajal.saha@unbc.ca, saikat.das@uvu.edu, gdecarvalho@brocku.ca}
       }

\maketitle

\begin{abstract}
In this paper, we present a novel hybrid deep learning model, named ConvLSTMTransNet, designed for time series prediction, with a specific application to internet traffic telemetry. This model integrates the strengths of Convolutional Neural Networks (CNNs), Long Short-Term Memory (LSTM) networks, and Transformer encoders to capture complex spatial-temporal relationships inherent in time series data. The ConvLSTMTransNet model was evaluated against three baseline models: RNN, LSTM, and Gated Recurrent Unit (GRU), using real internet traffic data sampled from high-speed ports on a provider edge router. Performance metrics such as Mean Absolute Error (MAE), Root Mean Squared Error (RMSE), and Weighted Absolute Percentage Error (WAPE) were used to assess each model's accuracy. Our findings demonstrate that ConvLSTMTransNet significantly outperforms the baseline models by approximately 10\% in terms of prediction accuracy. ConvLSTMTransNet surpasses traditional models due to its innovative architectural features, which enhance its ability to capture temporal dependencies and extract spatial features from internet traffic data. Overall, these findings underscore the importance of employing advanced architectures tailored to the complexities of internet traffic data for achieving more precise predictions.
\end{abstract}

\begin{IEEEkeywords}
Deep Learning, Internet traffic telemetry, CNN, LSTM, Transformer.
\end{IEEEkeywords}

\section{Introduction}
\label{introduction}

The exponential growth of Internet traffic, driven by both commercial applications and the increasing number of IoT devices, highlights the critical need for advanced traffic prediction mechanisms. While our study focuses on high-speed cellular networks, the insights and methodologies developed can significantly inform IoT network management. Accurate traffic forecasting is crucial for optimizing resource allocation, enhancing Quality of Service (QoS), and preventing network congestion across diverse network types \cite{wang2018iot}.

Deep Learning (DL), a subset of Artificial Intelligence (AI) inspired by the human brain, has proven particularly effective in modelling complex data patterns \cite{gupta2021artificial}. This capability is indispensable for handling the stochastic and dynamic nature of Internet traffic. Our research leverages DL to improve traffic prediction models in cellular networks, which share several underlying traffic characteristics and challenges with IoT networks, such as variability in traffic flow and the need for real-time data processing \cite{tahaei2020rise}.

Given the interconnectedness of modern digital ecosystems, techniques that improve forecasting accuracy in commercial networks also have potential applications in IoT scenarios. IoT networks, characterized by their vast number and diversity of connected devices, can benefit from the robust predictive models developed for high-volume cellular networks \cite{vaezi2022cellular}. These models can help manage the complexity and scale of IoT traffic, ensuring stability and efficiency.

In this paper, we propose ConvLSTMTransNet, a novel hybrid DL model that is meticulously designed for time series prediction, with a specific focus on Internet traffic telemetry—a domain characterized by dynamic and often unpredictable fluctuations. ConvLSTMTransNet represents a fusion of Convolutional Neural Networks (CNNs), Long Short-Term Memory (LSTM) networks, and Transformer encoders, which are jointly leveraged to capture the multifaceted relationships embedded within the Internet traffic time series data. This hybrid architecture enables ConvLSTMTransNet to effectively encode both spatial and temporal information, offering a comprehensive framework for forecasting tasks.

The performance of ConvLSTMTransNet is evaluated considering real Internet traffic data sampled from high-speed ports of a provider edge router and is compared against three baseline DL models - Recurrent Neural Network (RNN), LSTM, and Gated Recurrent Unit (GRU) - using the following metrics Mean Absolute Error (MAE), Root Mean Squared Error (RMSE), and Weighted Absolute Percentage Error (WAPE) are employed to quantitatively assess the accuracy and robustness of each model. Through a comprehensive experimental assessment, our findings demonstrate that ConvLSTMTransNet significantly outperforms the three baseline models under both scenarios. This capability underscores the potential of ConvLSTMTransNet to serve as a critical tool in both commercial and IoT network operations, supporting reliable and efficient network management. In summary, the main contributions of our work are as follows:

\begin{enumerate}
  
  \item We introduce ConvLSTMTransNet, a novel hybrid DL model meticulously designed for time series prediction, specifically tailored for Internet traffic telemetry. ConvLSTMTransNet represents a fusion of CNN, LSTM, and Transformer encoders, leveraging their respective strengths to capture the multifaceted relationships inherent in Internet traffic time series data.
  
  \item We conduct a thorough evaluation of ConvLSTMTransNet using real Internet traffic data sampled from high-speed ports of a provider edge router. We compare its performance against three baseline DL models - RNN, LSTM, and GRU - using key metrics such as MAE, RMSE, and WAPE.
  
  \item  Our experimental results demonstrate that ConvLSTMTransNet significantly outperforms the three baseline models under both scenarios. This superior performance underscores the potential of ConvLSTMTransNet to serve as a critical tool in both commercial and IoT network operations, supporting reliable and efficient network management.
  
  
\end{enumerate}

This paper is organized as follows. Section \ref{sec:literature} describes the literature review of current traffic prediction using machine learning models. Section \ref{sec:proposed methodology} presents our proposed methodology. Section \ref{sec:experiment} summarizes the experimentation configuration and discusses results for comparative analysis. Finally, section \ref{sec:conclusion} concludes our paper and sheds light on future research directions.

\section{Literature Review}
\label{sec:literature}

Network traffic prediction is a critical task in modern communication systems, influencing resource allocation, network management, network operation, and traffic optimization strategies. Recent advancements in DL have significantly enhanced the accuracy and efficiency of univariate time series prediction methods. This literature review examines cutting-edge approaches and the current state of the art in network traffic prediction, focusing on a hybrid DL approach for enhanced univariate time series prediction.

Jiang et al. \cite{jiang2022internet} highlighted the need for precise traffic prediction to enhance resource allocation and network management. They evaluated 13 deep neural networks and demonstrated their superiority over baseline models, particularly noting the effectiveness of InceptionTime. Abbasi et al. \cite{abbasi2021deep} focused on DL applications in network traffic monitoring and analysis, emphasizing their roles in handling complex traffic patterns. They reviewed DL techniques for Network Traffic Monitoring and Analysis (NTMA), discussing applications such as traffic classification and prediction. Lotfollahi et al. \cite{lotfollahi2020deep} proposed Deep Packet, a DL-based approach for encrypted traffic classification. Their method integrated feature extraction and classification, achieving high accuracy in traffic categorization and application identification tasks.

A comprehensive survey on traffic prediction using deep learning is essential for gaining insights into the existing research landscape and identifying critical gaps, facilitating the development of more robust and accurate prediction models tailored to real-world traffic dynamics. Ferreira et al. \cite{ferreira2023forecasting} presented a survey and tutorial on network traffic prediction techniques that shed light on autoregressive moving average models and artificial neural networks while conducting experimental assessment comparing their performance. Yin et al.\cite{yin2021deep} discussed the challenges in traffic prediction within intelligent transportation systems through the lenses of DL considering existing methods, real-world datasets, and performance. 

In addition to DL, general machine learning (ML)-based approaches, federated learning (FL), and CNNs are widely recognized and effectively utilized in traffic prediction tasks, offering diverse methodologies for modeling complex spatio-temporal dependencies in traffic data. Panayiotou et al. \cite{panayiotou2023survey} reviewed ML-based approaches, emphasizing their role in transitioning from reactive to adaptive optical networks. ML-aided service provisioning methods, including predictive and prescriptive frameworks, were discussed alongside their limitations and future opportunities. Sanon et al.\cite{sanon2023secure} evaluated homomorphic encrypted network traffic prediction using FL while addressing privacy concerns in sharing traffic information. They explored secure multi-party computation and presented an in-depth evaluation of the approach. Xu et al. \cite{xu2023adaptive} proposed AGFCRN, an adaptive graph fusion convolutional recurrent network for traffic forecasting. Their model dynamically captures temporal and spatial traffic characteristics, achieving competitive performance compared to state-of-the-art methods.

The combination of several DL methods has the potential to enhance predictive accuracy and robustness by leveraging complementary strengths and mitigating individual weaknesses which, ultimately, leads to solutions that outperform the standalone approaches. Wang et al. \cite{wang2023hybrid} introduced CEEMDAN-TGA, a hybrid DL method for network traffic prediction that combines Complete Ensemble Empirical Mode Decomposition with Adaptive Noise (CEEMDAN), Temporal Convolutional Network (TCN), Gated Recurrent Unit (GRU), and Attention Mechanism to improve prediction accuracy and stability, with promising results for network resource optimization and scheduling. Lai et al. \cite{lai2023deep} focused on traffic prediction for digital twin networks, proposing eConvLSTM, a DL-based method for accurate traffic synchronization. Their approach outperformed existing baselines, improving prediction accuracy while satisfying efficiency requirements for network resource planning and management. Miao et al. \cite{miao2023novel} proposed ESARIMA, a novel traffic prediction model combining autoregressive integrated moving average (ARIMA) with empirical mode decomposition (EMD) and singular value decomposition (SVD). Their method stabilized traffic data with EMD, reduced noise using SVD, and enhanced ARIMA efficiency and accuracy.

While recent studies have made significant improvements in network traffic prediction, there remains a notable gap in effectively integrating the strengths of different deep learning architectures to handle the multifaceted nature of network traffic data. Most current approaches tend to focus on either spatial or temporal aspects of the data but seldom both effectively. For instance, models like InceptionTime and AGFCRN, while robust, often prioritize temporal dynamics over spatial relationships or vice versa, potentially overlooking complex dependencies that could be pivotal for more accurate predictions. This paper narrows this critical gap in the literature by proposing ConvLSTMTransNet - a novel hybrid DL model that merges CNNs, LSTMs, and Multi-head Transformer architectures. Thanks to the ability to extract spatial features of CNNs, capture temporal dependencies of LSTM, and handle long-range interactions of Multi-head Transformers, ConvLSTMTransNet achieves high performance in terms of prediction accuracy. This innovative approach marks a significant step forward in the development of network traffic prediction models that are both highly accurate and robust while making ConvLSTMTransNet a compelling practical approach for Internet traffic telemetry.

\section{Proposed Methodology}
\label{sec:proposed methodology}

\begin{figure*}
    \centering
    \includegraphics[width=0.9\textwidth]{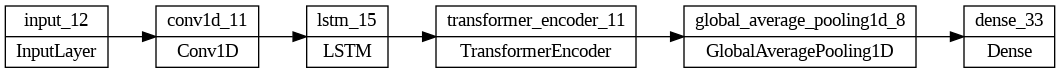}
    \caption{Proposed ConvLSTMTransNet Model Architecture Illustrating its Layers and Connections}
    \label{fig:model}
\end{figure*}

\subsection{Data Handling and Normalization}
An essential part of our data preprocessing pipeline involves handling missing values and the normalization of the dataset. These steps are fundamental to ensuring the integrity and uniformity of the data before it undergoes complex transformations within our predictive model.

Missing values in time series data can significantly impede the model's ability to learn accurate temporal dependencies. Such gaps disrupt the continuity essential for understanding patterns over time. To address this, we employ a forward-filling method (`fillna(method='ffill')`) where each missing value is replaced with the last observed value. This approach maintains the sequence's continuity, crucial for the model to accurately capture and predict future trends based on historical data. The forward-filling method is particularly suitable for time series data, where the assumption that changes between successive time points are incremental allows for a reasonable estimation of missing values. Given the diversity in scales across different features in time series datasets, normalization is a critical preprocessing step to ensure that our model can learn effectively. We utilize the MinMaxScaler to transform the feature of interest, in this case, 'bps', to a range between 0 and 1. This scaling not only helps in faster convergence during the training process but also prevents any single feature from disproportionately influencing the model's predictions due to scale differences. Normalizing the data to a common scale allows the model to treat all features equally, providing a balanced learning environment and improving the overall prediction accuracy.

\subsection{Time-Lagged Feature Extraction in Time Series Forecasting}
A critical step in our proposed model for time series forecasting involves the extraction of time-lagged features from the dataset. This process is foundational to capturing the temporal dependencies that characterize the sequential nature of time series data. Time-lagged features refer to previous time steps' data points used as input to forecast future values. The extraction of these features is pivotal for the model to learn from past trends and patterns to make accurate predictions.

Our approach involves segmenting the normalized time series data into sequences of a fixed length, referred to as \(seq\_len\). Each sequence serves as an input instance for the model, where the corresponding output is the value immediately following the sequence. This segmentation approach transforms the time series into a supervised learning problem, allowing the model to understand how past values influence future ones.

For a given time series, \(X = \{x_1, x_2, ..., x_n\}\), where \(x_i\) represents the data point at time \(i\), and a chosen sequence length of \(L\), we generate input sequences and their corresponding targets as follows: Each input sequence is formed by selecting \(L\) consecutive data points from the time series, \(S_i = \{x_i, x_{i+1}, ..., x_{i+L-1}\}\), with the target \(y_i = x_{i+L}\), representing the next value in the series. This process is repeated for all possible starting points in the time series, creating a dataset of input-target pairs \((S_i, y_i)\) for model training.



\subsection{Baseline Models}
To evaluate the performance of our proposed model, we selected three baseline models for comparison: RNN, LSTM networks, and GRU. These models are established in the literature for their efficacy in handling sequential data. Below, we detail each baseline model and provide a mathematical framework for their operations.

\subsubsection{Recurrent Neural Networks (RNN)}
They are foundational for processing sequential data, capturing temporal dependencies through hidden states passed through the sequence. For a given sequence $\{x_1, x_2, ..., x_T\}$, the RNN updates its hidden state $h_t$ at each time step $t$ using:

\[ h_t = \sigma(W_h h_{t-1} + W_x x_t + b) \]

\noindent where $W_h$ and $W_x$ are the weights for the hidden state and input, respectively, $b$ is a bias term, and $\sigma$ denotes an activation function, typically sigmoid or tanh. The final output $y_t$ at each time step can be calculated using the hidden state:

\[ y_t = W_y h_t + b_y \]

Despite their simplicity and power, RNNs are limited by the vanishing gradient problem, making learning long-term dependencies challenging.

\subsubsection{Long Short-Term Memory (LSTM)}
These networks address the limitations of traditional RNNs by introducing a memory cell and gates that regulate information flow. Each LSTM cell updates its state using the following equations:

\begin{itemize}
    \item Forget gate: $f_t = \sigma(W_f \cdot [h_{t-1}, x_t] + b_f)$
    \item Input gate: $i_t = \sigma(W_i \cdot [h_{t-1}, x_t] + b_i)$
    \item Cell update: $\tilde{C}_t = \tanh(W_C \cdot [h_{t-1}, x_t] + b_C)$
    \item New cell state: $C_t = f_t \ast C_{t-1} + i_t \ast \tilde{C}_t$
    \item Output gate: $o_t = \sigma(W_o \cdot [h_{t-1}, x_t] + b_o)$
    \item New hidden state: $h_t = o_t \ast \tanh(C_t)$
\end{itemize}

This architecture allows LSTMs to effectively learn from data points far back in the sequence.

\subsubsection{Gated Recurrent Units (GRU)}
It simplifies the LSTM architecture by combining the forget and input gates into a single update gate and merging the cell state and hidden state. The GRU operates using the following equations:

\begin{itemize}
    \item Update gate: $z_t = \sigma(W_z \cdot [h_{t-1}, x_t])$
    \item Reset gate: $r_t = \sigma(W_r \cdot [h_{t-1}, x_t])$
    \item Candidate activation: $\tilde{h}_t = \tanh(W \cdot [r_t \ast h_{t-1}, x_t])$
    \item New hidden state: $h_t = (1 - z_t) \ast h_{t-1} + z_t \ast \tilde{h}_t$
\end{itemize}

GRUs retain the LSTM's ability to mitigate the vanishing gradient problem while being computationally more efficient due to their simpler structure.

Each of these baseline models—RNN, LSTM, and GRU—has been instrumental in advancing the field of sequence modeling and time series analysis. Their mathematical frameworks support their capacity to learn temporal dependencies, though with varying degrees of complexity and effectiveness. Our comparative analysis aims to explain the strengths and weaknesses of our proposed model relative to these well-established benchmarks in handling time series forecasting tasks.

\subsection{Proposed ConvLSTMTransNet Model Architecture}

In this study, we propose a hybrid model for time series forecasting that combines the capabilities of CNNs, LSTM, and Transformer encoders depicted in Fig. \ref{fig:model}. Initially, the model employs a Conv1D layer to capture local patterns within the time series data, utilizing the ReLU activation function to introduce non-linearity. This is followed by an LSTM layer, adept at capturing temporal dependencies through its gating mechanisms, thus retaining crucial information over long sequences. The inclusion of a Transformer encoder layer results in a crucial enhancement, enabling the model to distinguish long-range dependencies within the data. This is achieved through a multi-head self-attention mechanism that dynamically weights the significance of different parts of the input sequence, a feature particularly beneficial for the complicated patterns often found in time series forecasting. Furthermore, the model architecture is efficient by the exclusion of a Transformer decoder, a decision emphasized by the model's focus on forecasting future values from past observations rather than translating between sequences. This exclusion simplifies the model while maintaining its predictive ability, as evidenced by the subsequent application of GlobalAveragePooling1D and a dense output layer, which together undertake the learned features to a predicted future value. The model is trained using the Adam optimizer and evaluated on its forecasting accuracy and robustness to adversarial examples, underscoring its effectiveness and resilience in handling time series data. By carefully architecting this model, we connect the distinct advantages of CNNs, LSTMs, and Transformer encoders, creating a robust forecasting tool proficient at learning the complexities of time series data without the computational overhead and architectural complexity of a full Transformer model.

\subsubsection{Input and Convolutional Layer}

The model starts with a \texttt{Conv1D} layer, which applies convolution operations on the input sequences. Each convolution operation involves sliding a kernel (or filter) across the input sequence and computing the dot product between the kernel and the input at each position. For a given input sequence $x$ and kernel $k$, the convolution ($c$) at each position $i$ is calculated as:

\[
c_i = \sum_{j=0}^{m-1} k_j \cdot x_{i+j}
\]

\noindent where $m$ is the kernel size and $j$ indexes the kernel. The \texttt{Conv1D} layer's role is to extract local patterns within the input sequence, with the ReLU activation function ($\text{ReLU}(x) = \max(0, x)$) introducing non-linearity into the model.

\subsubsection{LSTM Layer}
Following the convolutional layer, an LSTM layer processes the data sequentially. The LSTM updates its cell state $C_t$ and hidden state $h_t$ at each time step $t$ based on the previous hidden state $h_{t-1}$, the previous cell state $C_{t-1}$, and the current input $x_t$, through a series of gates: the forget gate $f_t$, input gate $i_t$, and output gate $o_t$.

\subsubsection{Transformer Encoder Layer}

The Transformer Encoder layer utilizes a multi-head self-attention mechanism and position-wise feed-forward networks.

\begin{itemize}
    \item \textbf{Multi-Head Attention}: This mechanism allows the model to focus on different positions of the input sequence for each head. The attention scores are computed using scaled dot-product attention:
    \[
    \text{Attention}(Q, K, V) = \text{softmax}\left(\frac{QK^T}{\sqrt{d_k}}\right)V
    \]
    \noindent where $Q$, $K$, and $V$ are the query, key, and value matrices derived from the input, and $d_k$ is the dimension of the key vectors. The output of the attention mechanism for each head is concatenated and then projected to the desired dimension.
    
    \item \textbf{Position-wise Feed-Forward Networks}: Each position in the encoder's output undergoes the same feed-forward network operation independently:
    \[
    \text{FFN}(x) = \max(0, xW_1 + b_1)W_2 + b_2
    \]
    \noindent where $W_1$, $W_2$, $b_1$, and $b_2$ are the weights and biases of the feed-forward network, and ReLU provides non-linearity.
    
    \item \textbf{Layer Normalization and Dropout}: To stabilize the learning process and prevent overfitting, layer normalization and dropout are applied. Layer normalization adjusts the activations within a layer to have a mean of 0 and a standard deviation of 1, while dropout randomly sets a fraction of input units to 0 at each update during training time.
\end{itemize}

\subsubsection{GlobalAveragePooling1D and Output Layer}

Finally, the \texttt{GlobalAveragePooling1D} layer reduces the dimensionality of the encoder output by computing the average over the temporal dimension

\section{Experiment and Result Analysis}
\label{sec:experiment}


\subsection{Dataset Description}
In this experiment, real internet traffic telemetry was utilized from several high-speed interfaces. The telemetry data was collected by sampling the SNMP (Simple Network Management Protocol) interface MIB (Management Information Base) counter on a core-facing interface of a provider edge router. Samples were taken every 5 minutes, with the bps (bits per second) value calculated as the difference between the samples at the start and end of each interval, multiplied by 8. The interface had a capacity of 40 Gbps, ensuring no discards occurred during the sampling period. Our dataset comprises 8,563 samples, representing 29 full days of data (288 samples per day), with the final day's data being incomplete. Only the timestamp (GMT) and traffic data from the original JSON (JavaScript Object Notation) file were retained, with all other information discarded. Consequently, 29 days of data were used to develop our prediction model.


\subsection{Evaluation Metrics}
\label{metric}
To assess the performance of our traffic forecasting models, we employed Mean Absolute Error (MAE), Root Mean Squared Error (RMSE), and Weighted Average Percentage Error (WAPE). MAE and RMSE provide insights into the absolute and squared deviations between the predicted and actual values, respectively, while WAPE offers a weighted perspective on the percentage deviation. 

\begin{equation}
    MAE = \frac{1}{n} \sum_{i=1}^n |p_i - o_i|
\end{equation}

\begin{equation}
    RMSE = \sqrt{\frac{1}{n} \sum_{i=1}^n (p_i - o_i)^2}
\end{equation}

\begin{equation}
    WAPE = \frac{\sum_{i=1}^n |p_i - o_i|}{\sum_{i=1}^n |o_i|} \times 100\%
\end{equation}

The metric calculations considers $p_i$ as the predicted value, $o_i$ as the original value, and $n$ as the total number of test instances.

\begin{table}
\centering
\caption{Comparative Performance Metrics of RNN, LSTM, GRU, and ConvLSTMTransNet Models with Varying Sliding Window Length of 6 and 12.}
\begin{tabular}{l|ccc|ccc} 
\hline\hline
                 & \multicolumn{3}{c|}{6 input} & \multicolumn{3}{c}{12 input}  \\ 
\hline\hline
                 & MAE  & RMSE & WAPE           & MAE  & RMSE & WAPE            \\ 
\hline
RNN              & 0.38 & 0.55 & 4.04           & 0.64 & 0.82 & 6.82            \\
LSTM             & 0.36 & 0.56 & 3.87           & 0.41 & 0.54 & 4.37            \\
GRU              & 0.64 & 0.86 & 6.81           & 0.46 & 0.61 & 4.87            \\
Proposed & 0.30 & 0.48 & 3.27           & 0.37 & 0.59 & 3.91            \\
\hline\hline
\end{tabular}
\end{table}

\subsection{Result and Discussion}
The comparative analysis of model performance is essential for evaluating their effectiveness in predicting internet traffic. By analyzing the data in the table, distinct trends become apparent across the models and varying input window lengths. Specifically, focusing on metrics like Mean Absolute Error (MAE), Root Mean Squared Error (RMSE), and Weighted Absolute Percentage Error (WAPE) allows us to discern each model's predictive accuracy.

Firstly, by focusing on MAE, RMSE, and WAPE, we can distinguish each model's predictive accuracy. The RNN and LSTM models generally perform similarly, with LSTM often slightly outperforming RNN across both 6 and 12-input window lengths. For instance, when considering a 6-input window length, RNN has an MAE of 0.38, RMSE of 0.55, and WAPE of 4.04, while LSTM exhibits slightly lower values with MAE of 0.36, RMSE of 0.56, and WAPE of 3.87. The data indicates LSTM shows a 5.56\% improvement in MAE compared to RNN, along with a 1.82\% improvement in RMSE and a 4.45\% improvement in WAPE. This trend suggests that LSTM, with its ability to remember longer-term dependencies, may offer a slight advantage over traditional RNN architectures in capturing the temporal dynamics of internet traffic data.

Contrastingly, the GRU model tends to exhibit higher error metrics compared to RNN and LSTM, indicating a lower predictive accuracy. For example, with a 6-input window length, GRU shows an MAE of 0.64, RMSE of 0.86, and WAPE of 6.81. It shows a 68.42\% increase in MAE, a 56.36\% increase in RMSE, and a 68.67\% increase in WAPE compared to LSTM, highlighting its inferior performance compared with LSTM.

\begin{figure}[ht!]
    \centering
        \centering
        \includegraphics[width=0.45\textwidth]{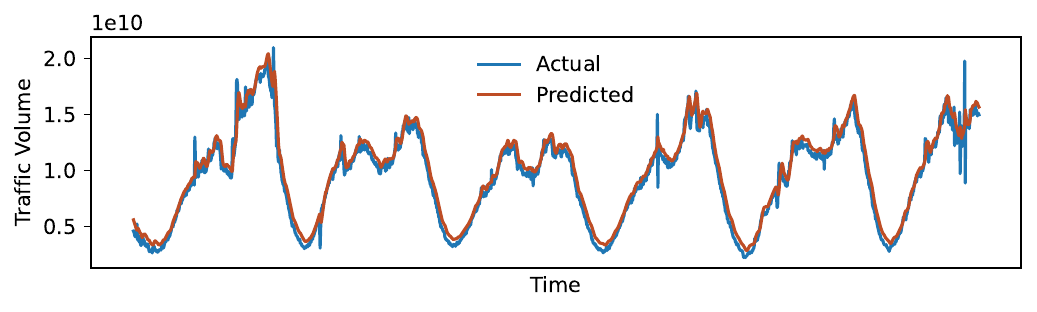}
        \label{fig:image1}
    

        
    
    \caption{Comparative Visualization of Actual vs. Predicted Traffic Using the ConvLSTMTransNet Model}
\label{fig:overall_traffic}
\end{figure}

However, the most striking result emerges from the performance of the proposed model. It consistently outperforms all other models across both input window lengths. For instance, with a 6-input window length, the proposed model achieves an MAE of 0.30, RMSE of 0.48, and WAPE of 3.27, showcasing its superior predictive capabilities compared to RNN, LSTM, and GRU. ConvLSTMTransNet achieves a 16.67\% improvement in MAE, a 12.50\% improvement in RMSE, and a 10.19\% improvement in WAPE compared to baseline models. We present the actual vs. predicted traffic by the best-performing ConvLSTMTransNet in Fig. \ref{fig:overall_traffic}.

The superior performance of the proposed ConvLSTMTransNet model compared to traditional models like RNN, LSTM, and GRU can be attributed to several key factors. Firstly, ConvLSTMTransNet incorporates innovative architectural features that enhance its ability to capture temporal dependencies in internet traffic data. By combining convolutional layers with LSTM or GRU layers, the model can extract spatial features from the input data before processing it temporally, resulting in improved predictive accuracy. Additionally, ConvLSTMTransNet excels at learning informative feature representations from the input data. Through hierarchical feature representation learning facilitated by convolutional layers, the model can automatically learn discriminative representations that contribute to better prediction of internet traffic dynamics. Moreover, the model is more robust to temporal variability in internet traffic data due to its ability to capture spatial patterns invariant to translation, rotation, and scaling. This robustness extends to temporal variability, enabling the model to generalize better to unseen temporal patterns.

In summary, while RNN and LSTM models offer competitive performance, with LSTM often exhibiting a slight edge, the GRU model lags behind in predictive accuracy. Conversely, the proposed model stands out as the most effective, demonstrating significantly lower error metrics and thus providing more precise predictions of internet traffic behavior. These findings underscore the importance of employing advanced architectures tailored to the complexities of internet traffic data for achieving superior predictive performance.

\section{Conclusion}
\label{sec:conclusion}

This paper introduced ConvLSTMTransNet, a novel hybrid model that adeptly incorporates the strengths of CNN, LSTM, and Transformer encoders to address the complexities of Internet traffic forecasting. Our comprehensive experiments, conducted using real internet traffic data, have validated the superior performance of ConvLSTMTransNet over standalone and traditional sequence modeling techniques such as RNN, LSTM, and GRU models in terms of accuracy. The ConvLSTMTransNet model's innovative architecture endows it with the ability to effectively encode spatial and temporal information, leading to enhanced forecasting precision. This was evidenced by its consistently lower MAE, RMSE, and WAPE scores across all tested scenarios. 

In future, we will extend ConvLSTMTransNet to multi-variate time series prediction to further enhance its utility. Additionally, adapting ConvLSTMTransNet to operate in an online learning mode could enable real-time traffic forecasting. Furthermore, we plan to investigate and enhance the model's robustness against adversarial examples generated through the Fast Gradient Sign Method (FGSM) approach.


\bibliographystyle{IEEEtran}
\bibliography{ref}

\begin{thebibliography}{10}
\providecommand{\url}[1]{#1}
\csname url@samestyle\endcsname
\providecommand{\newblock}{\relax}
\providecommand{\bibinfo}[2]{#2}
\providecommand{\BIBentrySTDinterwordspacing}{\spaceskip=0pt\relax}
\providecommand{\BIBentryALTinterwordstretchfactor}{4}
\providecommand{\BIBentryALTinterwordspacing}{\spaceskip=\fontdimen2\font plus
\BIBentryALTinterwordstretchfactor\fontdimen3\font minus \fontdimen4\font\relax}
\providecommand{\BIBforeignlanguage}[2]{{%
\expandafter\ifx\csname l@#1\endcsname\relax
\typeout{** WARNING: IEEEtran.bst: No hyphenation pattern has been}%
\typeout{** loaded for the language `#1'. Using the pattern for}%
\typeout{** the default language instead.}%
\else
\language=\csname l@#1\endcsname
\fi
#2}}
\providecommand{\BIBdecl}{\relax}
\BIBdecl

\bibitem{wang2018iot}
D.~Wang, D.~Chen, B.~Song, N.~Guizani, X.~Yu, and X.~Du, ``From iot to 5g i-iot: The next generation iot-based intelligent algorithms and 5g technologies,'' \emph{IEEE Communications Magazine}, vol.~56, no.~10, pp. 114--120, 2018.

\bibitem{gupta2021artificial}
R.~Gupta, D.~Srivastava, M.~Sahu, S.~Tiwari, R.~K. Ambasta, and P.~Kumar, ``Artificial intelligence to deep learning: machine intelligence approach for drug discovery,'' \emph{Molecular diversity}, vol.~25, pp. 1315--1360, 2021.

\bibitem{tahaei2020rise}
H.~Tahaei, F.~Afifi, A.~Asemi, F.~Zaki, and N.~B. Anuar, ``The rise of traffic classification in iot networks: A survey,'' \emph{Journal of Network and Computer Applications}, vol. 154, p. 102538, 2020.

\bibitem{vaezi2022cellular}
M.~Vaezi, A.~Azari, S.~R. Khosravirad, M.~Shirvanimoghaddam, M.~M. Azari, D.~Chasaki, and P.~Popovski, ``Cellular, wide-area, and non-terrestrial iot: A survey on 5g advances and the road toward 6g,'' \emph{IEEE Communications Surveys \& Tutorials}, vol.~24, no.~2, pp. 1117--1174, 2022.

\bibitem{jiang2022internet}
W.~Jiang, ``Internet traffic prediction with deep neural networks,'' \emph{Internet Technology Letters}, vol.~5, no.~2, p. e314, 2022.

\bibitem{abbasi2021deep}
M.~Abbasi, A.~Shahraki, and A.~Taherkordi, ``Deep learning for network traffic monitoring and analysis (ntma): A survey,'' \emph{Computer Communications}, vol. 170, pp. 19--41, 2021.

\bibitem{lotfollahi2020deep}
M.~Lotfollahi, M.~Jafari~Siavoshani, R.~Shirali Hossein~Zade, and M.~Saberian, ``Deep packet: A novel approach for encrypted traffic classification using deep learning,'' \emph{Soft Computing}, vol.~24, no.~3, pp. 1999--2012, 2020.

\bibitem{ferreira2023forecasting}
G.~O. Ferreira, C.~Ravazzi, F.~Dabbene, G.~C. Calafiore, and M.~Fiore, ``Forecasting network traffic: A survey and tutorial with open-source comparative evaluation,'' \emph{IEEE Access}, vol.~11, pp. 6018--6044, 2023.

\bibitem{yin2021deep}
X.~Yin, G.~Wu, J.~Wei, Y.~Shen, H.~Qi, and B.~Yin, ``Deep learning on traffic prediction: Methods, analysis, and future directions,'' \emph{IEEE Transactions on Intelligent Transportation Systems}, vol.~23, no.~6, pp. 4927--4943, 2021.

\bibitem{panayiotou2023survey}
T.~Panayiotou, M.~Michalopoulou, and G.~Ellinas, ``Survey on machine learning for traffic-driven service provisioning in optical networks,'' \emph{IEEE Communications Surveys \& Tutorials}, 2023.

\bibitem{sanon2023secure}
S.~P. Sanon, R.~Reddy, C.~Lipps, and H.~D. Schotten, ``Secure federated learning: An evaluation of homomorphic encrypted network traffic prediction,'' in \emph{2023 IEEE 20th Consumer Communications \& Networking Conference (CCNC)}.\hskip 1em plus 0.5em minus 0.4em\relax IEEE, 2023, pp. 1--6.

\bibitem{xu2023adaptive}
Y.~Xu, Y.~Lu, C.~Ji, and Q.~Zhang, ``Adaptive graph fusion convolutional recurrent network for traffic forecasting,'' \emph{IEEE Internet of Things Journal}, 2023.

\bibitem{wang2023hybrid}
D.~Wang, Y.-Y. Bao, and C.-M. Wang, ``A hybrid deep learning method based on ceemdan and attention mechanism for network traffic prediction,'' \emph{IEEE Access}, 2023.

\bibitem{lai2023deep}
J.~Lai, Z.~Chen, J.~Zhu, W.~Ma, L.~Gan, S.~Xie, and G.~Li, ``Deep learning based traffic prediction method for digital twin network,'' \emph{Cognitive Computation}, vol.~15, no.~5, pp. 1748--1766, 2023.

\bibitem{miao2023novel}
Y.~Miao, X.~Bai, Y.~Cao, Y.~Liu, F.~Dai, F.~Wang, L.~Qi, and W.~Dou, ``A novel short-term traffic prediction model based on svd and arima with blockchain in industrial internet of things,'' \emph{IEEE Internet of Things Journal}, 2023.

\end{thebibliography}

\vspace{12pt}

\end{document}